\documentclass[AMS,stixtwocol]{WileyNJD-v1}

\articletype{Original Article}%

\received{-}
\revised{-}
\accepted{-}

\journalname{.}
\journallogo{.}{.}
\jurl{.}

\begin{document}

\title{Cross-lingual Transfer Learning for Fake News Detector in a Low-Resource Language}

\author[1]{Sangdo Han}


\authormark{Sangdo Han \textsc{et al}}






\abstract[Abstract]{Development of methods to detect fake news (FN) in low-resource languages has been impeded by a lack of training data. In this study, we solve the problem by using only training data from a high-resource language. Our FN-detection system permitted this strategy by applying adversarial learning that transfers the detection knowledge through languages. To assist the knowledge transfer, our system judges the reliability of articles by exploiting source information, which is a cross-lingual feature that represents the credibility of the speaker. In experiments, our system got 3.71\% higher accuracy than a system that uses a machine-translated training dataset. In addition, our suggested cross-lingual feature exploitation for fake news detection improved accuracy by 3.03\%.}

\keywords{Fake News Detection, Source Information, Knowledge Transfer, Credibility of Article, Credibility of Speaker}

%


\maketitle

\section{Introduction}
Fake news (FN) is a serious social problem, because it aggravates social confusion and promotes conflict and antagonism. FN has endangered democracy \citep{grinberg2019fake} and has caused criminal cases such as Pizzagate\footnotemark. FN accounted for 25\% of news shared on Twitter during the 2016 US presidential election \citep{bovet2019influence}. FN is being generated and spread in real time, so it is virtually impossible for humans to catch individually. Therefore, the demand for a technology that automatically detects FN is increasing.

Recent work on FN detection (FND) has developed in at least three ways. First, some researchers detected FN by encoding articles with source information like authors and publishers \citep{popat2018declare, wang2017liar, sitaula2020credibility} because the credibility of source information could represent the credibility of the articles. Second, some studies \citep{wang2018eann, khattar2019mvae} encoded images in the articles to consider multi-modal features, because images can also be an important feature to estimate reliability. Third, some studies attempted to apply social context FND by encoding news articles with the social media posts that they belong to \citep{qian2018neural, volkova2017separating}. All of these approaches can predict FN, but they have difficulty detecting FN for low-resource languages because they lack training data.

FND has been studied for low-resource languages including Filipino \citep{cruz-etal-2020-localization}, Urdu \citep{amjad2020data}, Amharic \citep{gereme2021combating}, and Bangla \citep{hossain2020banfakenews}. The studies enabled detection of FN in low-resource languages, but they all used labeled training data in the target languages. This approach does not solve the major problem of the lack of training data.

To solve the problem, we introduce a system that two main features: 1) it uses adversarial learning to FND in a low-resource language, 2) it exploits a source information about the credibility of the speaker.

Adversarial learning can enable the system to detect FN by applying well-designed objective functions which can extract language-independent article features optimally. This approach avoids the limitation that is encountered when generating training data in a low-resource language from high-resource language training data by using machine translation, which is that the limited accuracy of the translation system becomes a bottleneck.

Identification of the credibility of the source information is a critical feature to detect FN, because some sources commonly publish reliable articles or utter reliable statements, and are therefore identifiable as 'reliable sources', whereas other sources have reliability that has not been established, or that has been demonstrated to be poor. To apply knowledge about this distinction, we exploited the identity of speakers as source information by encoding their credibility. To encode speakers, we generated our own source embedding from news articles, and expected that the embedding represents the credibility of speakers. This method uses the principle that “you shall know a speaker’s credibility by the company it keeps”, which follows Firth's \citep{firth1962synopsis} principle “you shall know a word by the company it keeps”. Our experiments verified that our source embedding represents speakers’ credibility effectively.

\footnotetext{\url{https://www.washingtonpost.com/local/public-safety/comet-pizza-gunman-to-appear-at-plea-deal-hearing-friday-morning/2017/03/23/e12c91ba-0986-11e7-b77c-0047d15a24e0_story.html}}

Given an input article, our system extracts source information, labels it with the given article, then classifies whether the article is fake or not. Our key contributions of this paper are:

\noindent{\textbf{Model}: We suggest the first FND system in which language transfer is applied without use of data that are labeled in the target language.}

\noindent{\textbf{Source Information}: We exploit source information and automatically extract it from the input article. We also suggest novel methods to encode source information.}

\noindent{\textbf{Dataset}: We collected a new FN dataset that is labeled by experts in the source language English, and in the target language Korean.}

\section{Related Work}

In this section, we introduce existing FND studies, FND systems that exploit source information, FND datasets for low-resource languages, and language transfer techniques for natural language processing (NLP).

\subsection{FND}
Until now, research on FND has exploited three major factors: knowledge, context, and style \citep{potthast2018stylometric}. Approaches that exploit knowledge \citep{ciampaglia2015computational} detect information inconsistencies between web documents and a constructed knowledge base of news articles; verification of claims by reference to knowledge can distinguish credibility clearly, but the knowledge bases must be updated in real time, and this task is expensive. Approaches that exploit context \citep{ma2016detecting,ma2017detect,ma2018rumor} analyze social network responses about a news article, especially the pattern in which news is spread through social media; these approaches show the highest accuracy in the FND task, but the social response takes time, so the detection of FN can be delayed. Approaches that exploit style \citep{popat2018declare, wang2017liar, sitaula2020credibility} attempt to encode the writing style or appearance of words in FN as a verification feature; this approach has many advantages when applied to the latest news, and obtains quick response; therefore, in this paper, our approach focuses on style exploitation.

\subsection{Source Information for FND}
When the source of an article is credible, the article is most likely credible. From the knowledge, some FND systems \citep{popat2018declare, wang2017liar, sitaula2020credibility} exploit the source information to detect FN. They commonly exploited it by encoding information about the authors or publishers; i.e., affiliation of the author, credibility history of the publisher, or the number of authors in the article. They taught us that source information is useful, but these approaches require human labor, including extracting sources from the articles \citep{wang2017liar} or encoding the information by applying human-generated labels \citep{popat2018declare, sitaula2020credibility}. We suggested a fully automatic approach including extracting and encoding source information.

\subsection{FND Dataset for Low-resource Languages}
Several FND datasets are available in low-resource languages \citep{cruz-etal-2020-localization,gereme2021combating,hossain2020banfakenews, amjad2020bend}. The datasets are commonly credibility-labeled by considering crawled news articles from web pages or Facebook posts. However, the labeling principles used to label the datasets are not precise or transparent. For example, some methods assume that news articles from non-credible sources are all fake \citep{hossain2020banfakenews, amjad2020bend}, whereas those from credible sources are all real \citep{cruz-etal-2020-localization}, but this absolute dichotomy is not reliably true. Another method \citep{gereme2021combating} uses hired journalists to label the news, but the reasons used for labeling were not provided. In this study, our dataset includes news articles that have been verified by experts from public fact-check services like Politifact\footnote{https://www.politifact.com/} and SNUFactCheck\footnote{https://factcheck.snu.ac.kr/}, which provide the reasoning.

\subsection{Language Transfer for NLP}
Language transfer is currently one of the most-intensively studies research areas in NLP \citep{ huang2019cross, zhou2019dual, zhou2018massively, kocmi2018trivial}. However, existing studies mostly consider tasks that are very distinct from the FND task; examples include named-entity recognition, machine translation, and dialog systems. Among the whole tasks, the task that is most similar to FND is sentiment analysis. ADAN \citep{chen2018adversarial}, a cross-lingual sentiment analyzer, uses adversarial training to transfer the knowledge of sentiment analysis from English to Chinese. ADAN does not require labeled data in the target language as training data, so we adopted the basic idea in our system. Our system is a cross-lingual FN detector that transfers the knowledge of FND from English to Korean. Especially, our system encodes source information of articles to exploit it as a cross-lingual feature and the encoding methods are specialized for FND. To the best of our knowledge, this is the first work that applies language transfer to FND without using target-language training data.

\section{Data}
In this study, we generated a new FND dataset in English and Korean. In this section, we introduce our FND datasets that are automatically generated.

We collected labeled data in three steps. First, we collected news articles from fact-checking web services like PolitiFact and SNUFactCheck. Second, we ensured that our collected articles are claim-related articles from sources provided in fact-check services. Third, we set up our task as a binary classification task from multi-class classification data. All steps follow previous research, as given below.

First, various FND datasets have been collected from fact-check services \citep{popat2018declare, wang2017liar, rashkin2017truth, shu2020fakenewsnet, vo2021hierarchical} because the services provide labels with trustworthy reasoning. Our labeled dataset is also collected from fact-checking services.

Second, collecting claim-related articles from fact-check services is a common way to collect fake news articles \citep{popat2018declare, shu2020fakenewsnet, vo2021hierarchical, popat2017truth}. Some of the news articles have been deleted, so some studies collected claim-related articles by using search engines. However, this collection process could include verification articles, instead of the source article. To keep the high purity of the dataset, we limited our source to only articles that are provided by fact-checking services.

Third, conversion of a multi-class dataset to a binary-class dataset is a common approach \citep{popat2018declare, rashkin2017truth, vo2021hierarchical} because the intention of FND also includes finding the boundary between fake and real news.

\subsection{Labeled Korean Dataset}
We generated a labeled Korean dataset (Table 1) by crawling the SNUFactCheck web service, which provides FN verification and reasoning. SNUFactCheck provides six types of verification information: Speaker, Claim, Label, Reasoning, Sources, and Claim Source. A \textbf{Speaker} is a person or organization who makes claims. A \textbf{Claim} is the main idea that the speaker asserts. A \textbf{Label} is the verification of the claim, as provided by a journalist, who is an expert in the service. \textbf{Reasoning} is a supporting article that verifies the label provided by the expert. \textbf{Sources} are reference articles, either news or other materials, which may include reasoning. A \textbf{Claim Source} is a news article that includes the claim. Our labeled Korean dataset is constructed by crawling claim sources and the label that belongs to each article. To decrease the number of credibility classes from five (true, mostly true, half-true, mostly false, and false) to two (real, fake), we regarded the first three as 'real', and the last two as 'fake'. Our constructed labeled Korean dataset includes 528 news articles from March 2017 to May 2020. We undersampled FN data to balance our dataset to 50:50.

\subsection{Labeled English Dataset}
To generate the labeled English dataset (Table 1), we crawled the PolitiFact site, which provides U.S. FN verification and reasoning. PolitiFact provides five types of verification information, similar to the types provided in SNUFactCheck, except for the claim source. PolitiFact does not include a claim source, so we should choose the most claim-related news article from sources by calculating the similarity between the claim and news articles. To calculate the similarity between claims and news articles, we computed the term frequency-inverse document frequency (TF-IDF) vector similarity between them. For the English data, we also reduced six class labels (true, mostly true, half-true, mostly-false, false, and pants-on-fire) to two class labels by grouping the first three classes as 'real', and the last three as 'fake'. The labeled English dataset contains 12,547 news articles published from October 2009 to May 2020. We maintained a real-to-fake ratio of 50:50 for our training data. 

\begin{table}[ht]
\begin{center}
\caption{Statistics of labeled dataset.}
  \begin{tabular}{c c c c c c} 
 \hline
 Language & Fake & Real & Train & Valid & Test \\
 \hline
 Korean & 264 & 264 & 0 & 264 & 264 \\ 
 English & 6480 & 6467 & 10130 & 1247 & 1170\\ 
 \hline
\end{tabular}
\end{center}
\end{table}

\subsection{Unlabeled Korean Dataset}
Language transfer using adversarial learning requires a large set of data in a target language. To meet this requirement, we crawled 24,000 Korean news articles from a Korean news platform, Naver news\footnote{\url{https://news.naver.com/}}, that were published in the same period as the labeled English dataset. To minimize cultural differences between unlabeled Korean data and labeled English data, the unlabeled Korean data included international news only.

 \section{Method}
Our system\footnote{Our code is available at \url{https://github.com/hansd410/cross-lingFND}} (Figure 1) has three major components: a news encoder, a language discriminator, and an FN detector. The news encoder encodes input news articles with source information. The language discriminator identifies whether the news article is written in English or Korean. The FN detector determines whether the input article is fake or real.

\begin{figure}[!ht]
\centering
\includegraphics[width=\columnwidth]{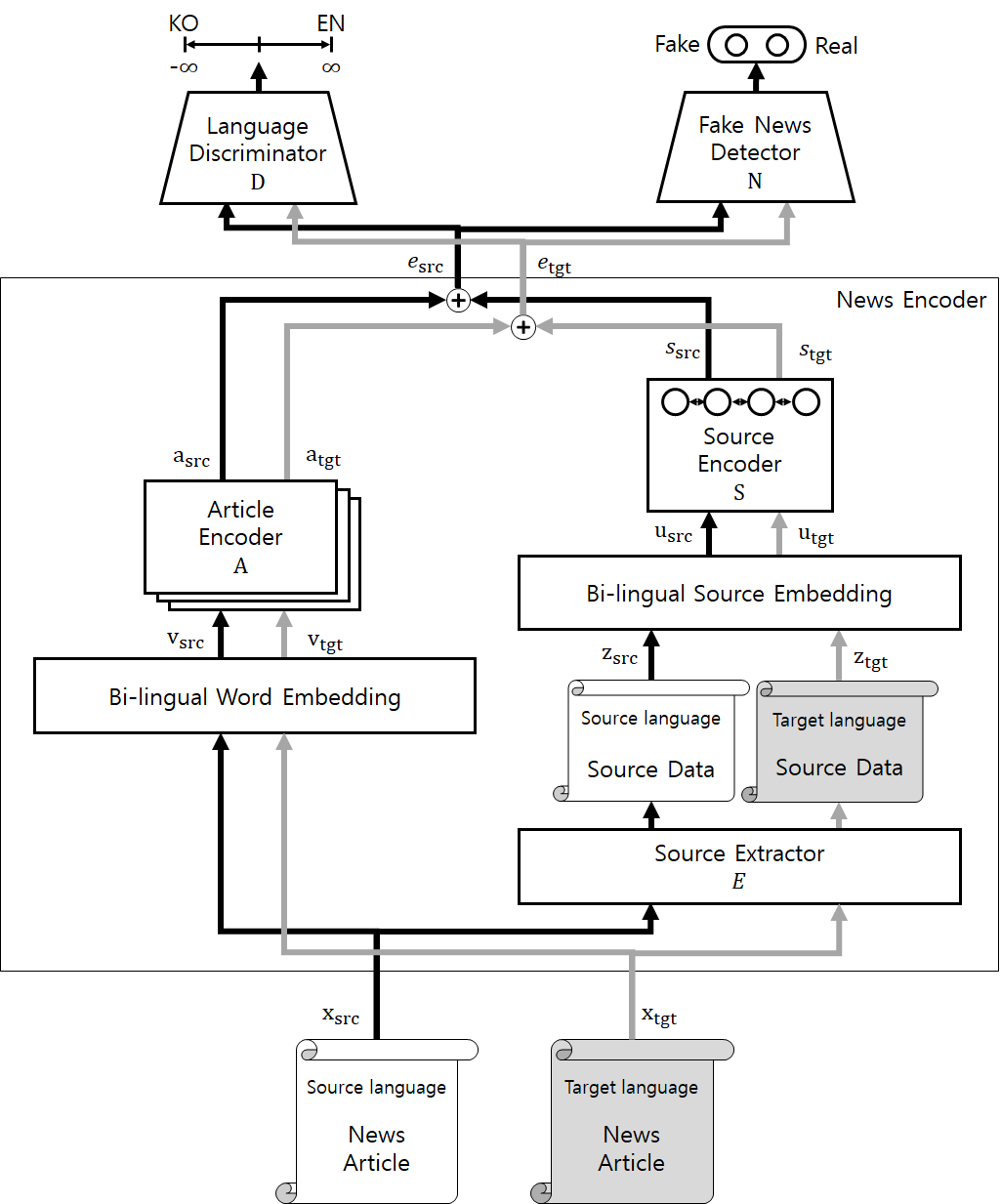}
\caption{Overall architecture. $x$: news article, $v$: article embedding matrix, $a$: article embedding vector, $z$: source data, $u$: source embedding matrix, $s$: source embedding vector, $e$: news embedding vector. Processes are described in the text.}
\end{figure}

\subsection{Overview}

Our system basically follows the architecture of ADAN \citep{chen2018adversarial}, which is a bilingual sentiment analyzer. The key differences between ADAN and our system are that our system includes a source encoder, a bilingual word embedding, and the output of news encoder.

The encoder includes two major components: article encoder $A$ and source encoder $S$. $A$ generates an article-embedding vector $a$ from an article representation matrix $v$ which is transformed from an input article $x$ by using bilingual word embedding (BWE). We generated our own BWE by aligning pre-trained 100-dimensional Korean\footnote{\url{https://ratsgo.github.io/embedding/}} and English GloVe embeddings \citep{pennington2014glove} by applying Procrustes \citep{grave2019unsupervised} alignment. The alignment used a bilingual word dictionary given in \citep{lample2018word}. Our $A$ uses a convolutional neural network (CNN), which is commonly used to encode documents \citep{kim-2014-convolutional, conneau2017very}.

$S$ encodes the source information that is extracted from news articles. In this study, we exploited $S$ as a cross-lingual feature to assist the knowledge transfer by encoding the source information to its credibility. In general, source information includes press companies, authors, and speakers who make claims in news articles \citep{popat2018declare, wang2017liar, sitaula2020credibility}. We defined our source information as speakers who make claims in articles, because it is the most content-related information among these options. To fulfill the objective, the system generates source data by extracting speaker candidates from an input article, then classifies the candidates according to their credibility. 

An output of the news encoder from $x$ is a news-embedding vector $e$, which is generated by concatenating $a$ and $s$. The FN detector $N$ and the language discriminator $D$ make final decisions by considering $e$. $D$ discriminates the language of the article, and $N$ detects whether the article is fake or not. Both modules are feed-forward networks, but only $N$ has a softmax layer at the end because the output value of $D$ is assumed as Wasserstein distance between the distributions of $e$ in both languages.

\subsection{Source Extractor}

To exploit source information, the source extractor $E$ generates source data $z$ from $x$. $z$ is composed of speakers who have made a claim in news articles, and is obtained by extracting speaker candidates from $x$. Speaker candidates are extracted from $x$ by recognizing PERSON-typed named entities, by using the SpaCy\footnote{\url{https://spacy.io/}} python library for English and named-entity extractor for Korean\footnote{\url{https://github.com/eagle705/pytorch-bert-crf-ner}}. The extractor generates source data by arranging the candidates in appearance order.

Our extracting strategy uses analysis of speakers in news articles. First, we identified the speakers: from the page that provides a whole list of speakers in PolitiFact\footnote{\url{https://www.politifact.com/personalities/}}, we crawled and analyzed all of their labels. They show the identity of each speaker, including political parties and jobs. We found that speakers are mostly people, especially politicians (Figure 2). In detail, the ‘None’ class is large, so we analyzed 100 random samples from it; 52 were people and 48 were organizations.

\begin{figure}[h]
\centering
\includegraphics[width=\columnwidth]{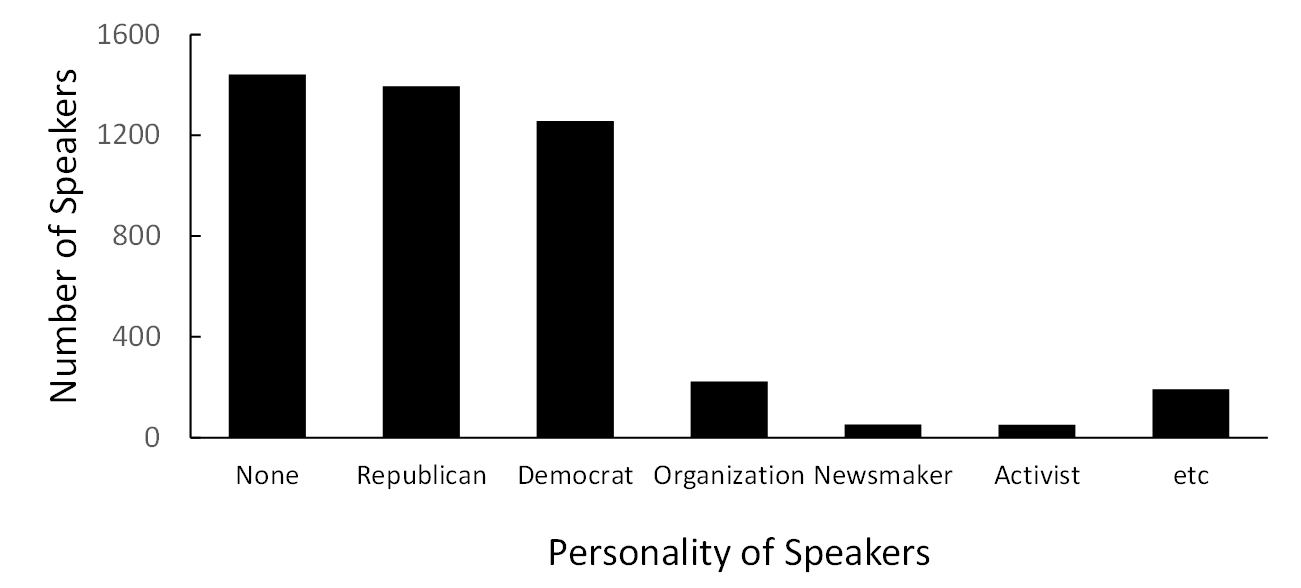}
\caption{Number of speakers by affiliation.}
\end{figure}

Second, to identify where the speakers appear in the news articles, we used a percentage to represent the location of the index of speakers relative to the beginning and end of the article. Most speakers were identified near the beginning of news articles (Figure 3).

\begin{figure}[h]
\centering
\includegraphics[width=\columnwidth]{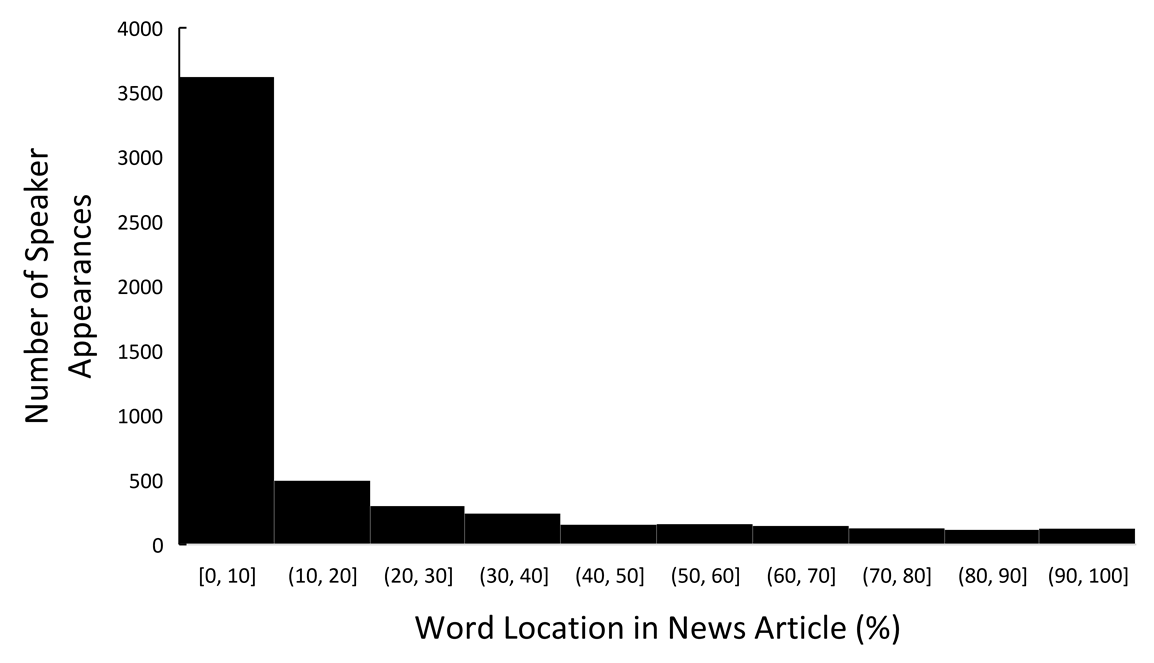}
\caption{Location of speakers in news articles.}
\end{figure}

These analyses yielded two major insights. First, we can collect speaker candidates by extracting the names of people and organizations from news articles. Second, the sequence of speaker candidates is informative because a candidate that is mentioned early is usually the speaker. From the additional analysis (Section 5.2), we determined that identifying PERSON-typed named entities in appearance order is the best way to generate the source data.

\begin{figure*}[h]
\centering
\includegraphics[width=\textwidth]{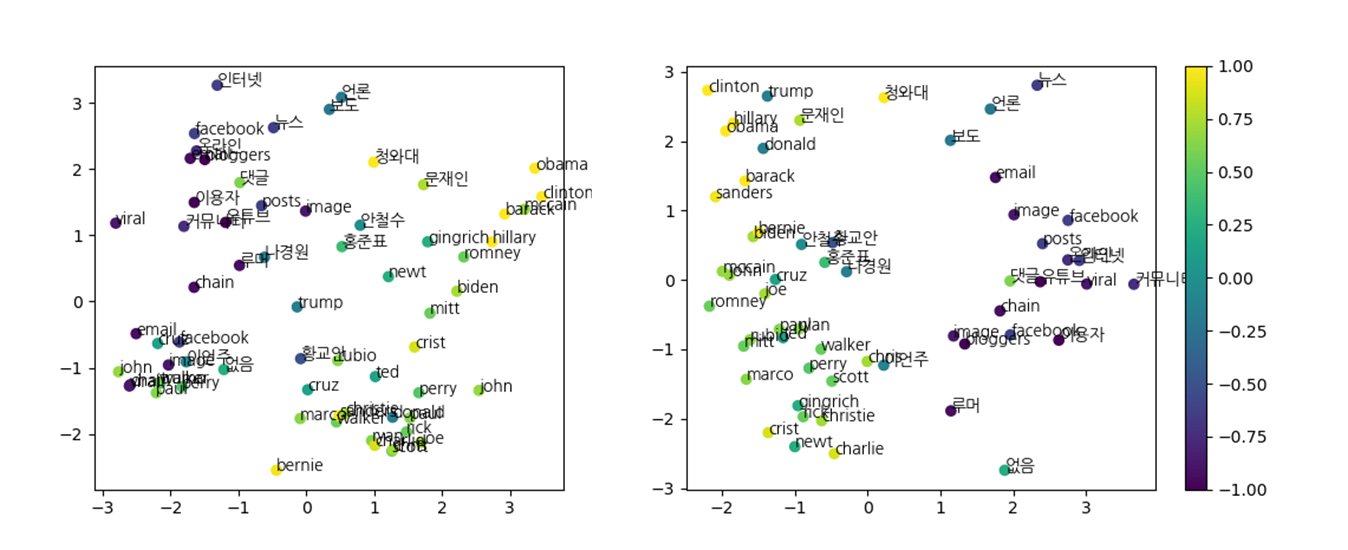}
\caption{Plotted embedding graph with colored credibility score. a) plot of bilingual word embedding, b) plot of bilingual source embedding. Scale bar: yellow = credible, blue = non-credible.}
\end{figure*}

\subsection{Bilingual Source Embedding}

To encode $z$ as a cross-lingual feature for FND, a word embedding that represents speakers’ credibility is required. In addition, the word embedding must be bilingual to exploit source data as a cross-lingual feature. To fulfill those requirements, we generated a new bilingual word embedding, bilingual source embedding (BSE).

To represent the credibility of speakers when they make claims, we generated BSE from news articles published in the same period as the labeled dataset. For Korean, we generated embedding space by analyzing 33,000 crawled domestic Korean news articles that were published in the same period as the labeled Korean dataset. For English, we used pre-existing labeled data. Our idea to generate BSE applies the intuition that speakers who are not credible will produce fake news more frequently than real news. After generating two source embeddings in each language, we convert them to a bi-lingual embedding by applying Procrustes \citep{grave2019unsupervised} alignment.

To verify whether the source embedding represents credibility, we plotted the top twenty most frequently appearing speakers of our labeled dataset in both languages in a two-dimensional space (Figure 4), then colored each speaker according to the credibility score that was calculated from our labeled dataset. Calculation of credibility score $\hat{s_i}$ for speaker $i$ is 

\[
b_{ij}=
\begin{cases}
-1,&\text{fake news}\\
1,&\text{real news}
\end{cases}
\]
\[
s_i = \frac{\sum_{j=1}^{c_i}b_{ij}}{c_i}
\]
\begin{equation}
\hat{s_i}=\frac{s_i-min(s)}{max(s)-min(s)}
\end{equation}
where $c_i$ represents the number of articles in which speaker $i$ appeared with score $b_{ij}$ for each article $j$. BSE space (Figure 4b) shows clusters divided according to credibility, whereas BWE space (Figure 4a) assembles speakers that have different credibility scores. For example, Figure 4a reveals clusters of speakers that have different credibility scores around (-2,-1), (1,-2) coordinates, but these clusters are not apparent in Figure 4b.

As a quantitative measure, we calculated credibility score difference $diff_{cred}$ between nearby speakers. For $m$ speakers with $n$ nearest neighbors, $diff_{cred}$ is calculated as

\begin{equation}
diff_{cred}=\frac{\sum_{i=1}^{m}\sum_{j=1}^{n}|\hat{s_i}-\widehat{s_{N_{ij}}}|}{m*n}
\end{equation}
where $N_{ij}$ represents the $j$th nearest neighbor of speaker $i$. Our calculation used $m$ = 40 and $n$ = 5. BWE achieved $diff_{cred}$ = 0.4171, and BSE achieved $diff_{cred}$ = 0.2789. As we expected, BSE got lower speaker distance than BWE.

\subsection{Source Encoder}

After the BSE generates an embedding matrix $u$, $S$ encodes it to a vector $s$ that records all source information of news article $x$. $S$ encodes $u$ on bi-directional LSTM (bi-LSTM) to encode sequential information together. The average of bi-LSTM nodes becomes $s$, which was the best encoding option from Section 5.2.

\subsection{Training Details}

Our system also follows training procedure of a Wasserstein generative adversarial network (WGAN) \citep{martin2017wasserstein}, as ADAN did. To exploit source information for FND, the objective functions are modified. Especially, the language discriminator uses

\begin{align}
\begin{split}
J\textsubscript{d}(\theta\textsubscript{a},\theta\textsubscript{s})
\equiv
\max\limits_{{\theta\textsubscript{d}}}
E\textsubscript{$x_{src}\sim X_{src}$}[D(a_{src};s_{src})] \\
- E\textsubscript{$x_{tgt}\sim X_{tgt}$}[D(a_{tgt};s_{tgt})], 
\end{split}
\end{align}
the FN detector uses
\begin{equation}
J\textsubscript{n}(\theta\textsubscript{a},\theta\textsubscript{s})
\equiv
\min\limits_{{\theta\textsubscript{n}}}
E\textsubscript{(x,y)}[L_n(N(a;s),y)] , 
\end{equation}
the article encoder uses
\begin{equation}
J\textsubscript{a}
\equiv
\min\limits_{{\theta\textsubscript{a}}}
J_d(\theta_a)+\lambda J_n(\theta_a), 
\end{equation}
and the source encoder uses
\begin{equation}
J\textsubscript{s}
\equiv
\min\limits_{{\theta\textsubscript{s}}}
J_d(\theta_s)+\lambda J_n(\theta_s),
\end{equation}
where $\theta_d$ are parameters of language discriminator $D$, $\theta_n$ are parameters of FN detector $N$, $\theta_a$ are parameters of article encoder $A$, and $\theta_s$ are parameters of source encoder $S$. From dataset $X$, $x$ represents sampled data, $y$ includes labels of $x$, $a$ represents encoded vector from article encoder, and $s$ is the encoded vector from source encoder. As in ADAN, we assumed Eq. (1) as Wasserstein distance between distributions of source and target language data. As we assumed, the neural net tries to increase $J_d$ for English articles and to decrease it for Korean articles. $J_n$ minimizes cross-entropy loss $L_n$ between the FND result and the label $y$. Finally, $A$ and $S$ minimize both objectives, $J_d$ and $J_n$, to discourage $D$ and encourage $N$ simultaneously. In this work, we set hyperparameters $\lambda$ = 0.01, $c$ = 0.01 and $k$ = 5.

Hyperparameters of our system are either pre-existing or newly-added. Values of the newly-added parameters were selected by considering pre-defined parameters as given below. $A$ has 3-, 4- and 5-sized 400 kernels for each size and $a$ is a 900-dimensional vector. $S$ encodes $s$ as a 900-dimensional vector by applying BSE with bi-LSTM. After applying source information, $N$ and $D$ got an 1,800-dimensional vector as input. $N$ and $D$ are binary classifiers that use fully-connected networks with two hidden layers, including 1,800 hidden units, and only $N$ includes ReLUs to resolve non-linearities. For training, we applied Adam \citep{kingma2015adam} optimizer with 0.0005 learning rate for each module.

We trained our system for 30 epochs on the validation set of a labeled Korean dataset and used early stopping to select the best model.

Our training process uses Nvidia GeForce TitanX, which has GDDR 5 12GB memory and a 1,000-Mhz GPU base clock. The whole training process took 20 hours.

\section{Experiment}

\subsection{Overall}
In experiments, we compared our system with three baselines: ADAN, a system that uses machine translation (SMT), and text-CNN \citep{kim-2014-convolutional} (Table 2). By comparing with ADAN, we can verify that exploitation of source information is crucial for cross-lingual FND. SMT uses machine translation without language transfer; this system includes $A$ and $N$ only, and is trained on a labeled English dataset and tested on a labeled Korean dataset that had been translated to English by Google translate\footnote{\url{https://cloud.google.com/translate}}. Text-CNN is an effective text encoder for FND \citep{wang2018eann}. No labeled training dataset is available for Korean, so we measured Text-CNN accuracy by five-fold cross-validation of a labeled Korean dataset. We also added RANDOM accuracy that randomly selects FN from input articles for comparison at the end. Our system achieved the highest accuracy, especially the highest F1 score among the methods. All given experimental results are averages of five trials with a 200-word length limitation for calculation efficiency.

\begin{table}[h]
\begin{center}
\caption{Overall accuracy of FND in Korean.}
 \begin{tabular}{c c c c c} 
 \hline
 System & Accuracy [\%] & Precision & Recall & F1 \\
 \hline
 Proposed & 57.35 & 0.5712 & 0.5985 & 0.5789 \\ 
 ADAN & 54.32 & 0.5446 & 0.4636 & 0.4880 \\
 SMT & 53.64 & 0.6030 & 0.2667 & 0.3325 \\
 Text-CNN & 51.44 & 0.5136 & 0.5545 & 0.5328 \\
 RANDOM & 49.62 & 0.4962 & 0.4924 & 0.4943 \\
 \hline
\end{tabular}
\end{center}
\end{table}

Results (Table 2) demonstrated that a cross-lingual approach that exploits source information is effective for FND in a low-resource language. In detail, compared with ADAN (Table 2, row 2), our system is well-improved for FND by exploiting source information as a cross-lingual feature. SMT (Table 2, row 3) got similar accuracy to ADAN, but with very low recall; i.e., the detection result of the system is biased toward real news. We conclude the reason that it is not trained well because the quality of machine-translated articles is insufficient. In the case of Text-CNN (Table 2, row 4), the FND accuracy with a small amount of training data in the target language was similar to RANDOM (Table 2, row 5). We conclude that language transfer for FND in a low-resource language is essential.

\begin{table*}[h]
\centering
\caption{Analysis of source information components. PER: person; ORG: organization.}
 \begin{tabular}{c l l c c c c} 
 \hline
 Row & Component & Features & Accuracy [\%] & Precision & Recall & F1 \\ 
 \hline
 1 &  Source Data & PER \& ORG & 54.47 & 0.6125 & 0.3712 & 0.3722 \\
2 &   & ORG & 53.94 & 0.5521 & 0.4773 & 0.4965 \\
3 &   & PER (shuffled) & 53.18 & 0.5344 & 0.4924 & 0.4981 \\ 
4 &  Source Embedding & BWE & 55.76 & 0.5778 & 0.5000 & 0.5052 \\
5 & Source Encoder  & CNN & 51.59 & 0.5778 & 0.6758 & 0.5812 \\
6 &   & LSTM (last) & 56.14 & 0.5719 & 0.5697 & 0.5136 \\ 
7 &   & LSTM (first) & 53.26 & 0.5962 & 0.3727 & 0.3861 \\
8 &   & LSTM (attention) & 54.02 & 0.5582 & 0.4121 & 0.4571 \\

 \hline
\end{tabular}
\end{table*}

\subsection{Source Information Analysis}
In additional experiments, we analyzed the effect of components that exploit source information (Table 3). The analyzed components include source data, embedding, and encoder.

In Figure 2, most of the speakers are PERSON-type or ORGANIZATION-type named entities. To determine which type of named entity is essential, our experiment compared various source data by combining named entity types. We determined that PERSON is the most critical feature for transfer learning, whereas ORGANIZATION decreases the accuracy of FND (Table 3, row 1). We hypothesize that the accuracy is decreased by including ORGANIZATION type because organizations are so numerous that they do not provide reliable clues to detect FN. From the whole analysis, our final source data were generated by extracting PERSON-type named entities from a news article. We also checked whether the sequential information of source data is necessary, by shuffling speaker candidates to remove sequential information. This experiment demonstrated that to detect FN, sequential information of source data is necessary (Table 3, row 3).

To check the effect of BSE, we compared FND accuracy with source data encoded using BWE. Applying BSE got higher accuracy than BWE (Table 3, row 4). Interestingly, even BWE got relatively high accuracy; we suppose that this result occurs because some non-credible speakers who use social networking services are already separated from credible speakers in BWE space.

Finally, to find the best method of encoding the data, we compared CNN (Table 3, row 5), and other LSTM encoding methods: last node (Table 3, row 6), first node (Table 3, row 7), and attention-based \citep{luong2015effective} (Table 3, row 8). Our proposed system that averages hidden nodes of bi-LSTM, was more accurate than other encoding methods, and CNN was the least accurate.

\section{Conclusion}
This paper describes a system that can detect FN written in a low-resource language by transferring the knowledge from a high-resource language to the low-resource language. We also proposed a cross-lingual feature, i.e., source information, to assist the knowledge transfer without any human labor. Our system achieved higher accuracy of FND than baseline systems, and the source information increased the accuracy. Especially, the accuracy was increased by each component of source information, which includes source data, source embedding, and source encoder. Knowledge transfer and source information are easy to apply, so our approach will be useful for FND in other low-resource languages.

For future work, we plan to design an explainable FND system in low-resource languages. Explainability is one of the most necessary attributes for neural-network research. Especially, in FND, explainability is crucial because people want to understand the reason for an article’s lack of credibility. We plan to identify the feature or features that made the article be classified as FN. Critical features for cross-lingual fake news detection could give us new knowledge that the article includes without external documents. For example, applying explainability in cross-lingual FND could find critical clues (suspicious topics, writing styles, and speakers) to detect FN in the target language.


\nocite{*}
\bibliography{ETRIJ-v1}%

\providecommand{\bysame}{\leavevmode\hbox to3em{\hrulefill}\thinspace}
\providecommand{\MR}{\relax\ifhmode\unskip\space\fi MR }
\providecommand{\MRhref}[2]{%
  \href{http://www.ams.org/mathscinet-getitem?mr=#1}{#2}
}
\providecommand{\href}[2]{#2}
\begin{thebibliography}{10}

\bibitem{grinberg2019fake}
Nir Grinberg, Kenneth Joseph, Lisa Friedland, Briony Swire-Thompson, and David
  Lazer, \emph{Fake news on twitter during the 2016 us presidential election},
  Science \textbf{363} (2019), no.~6425, 374--378.

\bibitem{bovet2019influence}
Alexandre Bovet and Hern{\'a}n~A Makse, \emph{Influence of fake news in twitter
  during the 2016 us presidential election}, Nature communications \textbf{10}
  (2019), no.~1, 1--14.

\bibitem{popat2018declare}
Kashyap Popat, Subhabrata Mukherjee, Andrew Yates, and Gerhard Weikum,
  \emph{Declare: Debunking fake news and false claims using evidence-aware deep
  learning}, Proceedings of the 2018 Conference on Empirical Methods in Natural
  Language Processing, 2018, pp.~22--32.

\bibitem{wang2017liar}
William~Yang Wang, \emph{“liar, liar pants on fire”: A new benchmark
  dataset for fake news detection}, Proceedings of the 55th Annual Meeting of
  the Association for Computational Linguistics (Volume 2: Short Papers), 2017,
  pp.~422--426.

\bibitem{sitaula2020credibility}
Niraj Sitaula, Chilukuri~K Mohan, Jennifer Grygiel, Xinyi Zhou, and Reza
  Zafarani, \emph{Credibility-based fake news detection}, Disinformation,
  Misinformation, and Fake News in Social Media, Springer, 2020, pp.~163--182.

\bibitem{wang2018eann}
Yaqing Wang, Fenglong Ma, Zhiwei Jin, Ye~Yuan, Guangxu Xun, Kishlay Jha, Lu~Su,
  and Jing Gao, \emph{Eann: Event adversarial neural networks for multi-modal
  fake news detection}, Proceedings of the 24th acm sigkdd international
  conference on knowledge discovery \& data mining, 2018, pp.~849--857.

\bibitem{khattar2019mvae}
Dhruv Khattar, Jaipal~Singh Goud, Manish Gupta, and Vasudeva Varma, \emph{Mvae:
  Multimodal variational autoencoder for fake news detection}, The world wide
  web conference, 2019, pp.~2915--2921.

\bibitem{qian2018neural}
Feng Qian, Chengyue Gong, Karishma Sharma, and Yan Liu, \emph{Neural user
  response generator: Fake news detection with collective user intelligence.},
  IJCAI, vol.~18, 2018, pp.~3834--3840.

\bibitem{volkova2017separating}
Svitlana Volkova, Kyle Shaffer, Jin~Yea Jang, and Nathan Hodas,
  \emph{Separating facts from fiction: Linguistic models to classify suspicious
  and trusted news posts on twitter}, Proceedings of the 55th Annual Meeting of
  the Association for Computational Linguistics (Volume 2: Short Papers), 2017,
  pp.~647--653.

\bibitem{cruz-etal-2020-localization}
Jan Christian~Blaise Cruz, Julianne~Agatha Tan, and Charibeth Cheng,
  \emph{Localization of fake news detection via multitask transfer learning},
  Proceedings of the 12th Language Resources and Evaluation Conference
  (Marseille, France), European Language Resources Association, May 2020,
  pp.~2596--2604 (English).

\bibitem{amjad2020data}
Maaz Amjad, Grigori Sidorov, and Alisa Zhila, \emph{Data augmentation using
  machine translation for fake news detection in the urdu language},
  Proceedings of The 12th Language Resources and Evaluation Conference, 2020,
  pp.~2537--2542.

\bibitem{gereme2021combating}
Fantahun Gereme, William Zhu, Tewodros Ayall, and Dagmawi Alemu,
  \emph{Combating fake news in “low-resource” languages: Amharic fake news
  detection accompanied by resource crafting}, Information \textbf{12} (2021),
  no.~1, 20.

\bibitem{hossain2020banfakenews}
Md~Zobaer Hossain, Md~Ashraful Rahman, Md~Saiful Islam, and Sudipta Kar,
  \emph{Banfakenews: A dataset for detecting fake news in bangla}, Proceedings
  of the 12th Language Resources and Evaluation Conference, 2020,
  pp.~2862--2871.

\bibitem{firth1962synopsis}
JR~Firth, \emph{A synopsis of linguistic theory, 1930--1935; studies in
  linguistic analysis}, London: Philosophical Society (1962).

\bibitem{potthast2018stylometric}
Martin Potthast, Johannes Kiesel, Kevin Reinartz, Janek Bevendorff, and Benno
  Stein, \emph{A stylometric inquiry into hyperpartisan and fake news},
  Proceedings of the 56th Annual Meeting of the Association for Computational
  Linguistics (Volume 1: Long Papers), 2018, pp.~231--240.

\bibitem{ciampaglia2015computational}
Giovanni~Luca Ciampaglia, Prashant Shiralkar, Luis~M Rocha, Johan Bollen,
  Filippo Menczer, and Alessandro Flammini, \emph{Computational fact checking
  from knowledge networks}, PloS one \textbf{10} (2015), no.~6, e0128193.

\bibitem{ma2016detecting}
Jing Ma, Wei Gao, Prasenjit Mitra, Sejeong Kwon, Bernard~J Jansen, Kam-Fai
  Wong, and Meeyoung Cha, \emph{Detecting rumors from microblogs with recurrent
  neural networks},  (2016).

\bibitem{ma2017detect}
Jing Ma, Wei Gao, and Kam-Fai Wong, \emph{Detect rumors in microblog posts
  using propagation structure via kernel learning}, Association for
  Computational Linguistics, 2017.

\bibitem{ma2018rumor}
\bysame, \emph{Rumor detection on twitter with tree-structured recursive neural
  networks}, Association for Computational Linguistics, 2018.

\bibitem{amjad2020bend}
Maaz Amjad, Grigori Sidorov, Alisa Zhila, Helena G{\'o}mez-Adorno, Ilia
  Voronkov, and Alexander Gelbukh, \emph{“bend the truth”: Benchmark
  dataset for fake news detection in urdu language and its evaluation}, Journal
  of Intelligent \& Fuzzy Systems \textbf{39} (2020), no.~2, 2457--2469.

\bibitem{huang2019cross}
Lifu Huang, Heng Ji, and Jonathan May, \emph{Cross-lingual multi-level
  adversarial transfer to enhance low-resource name tagging}, Proceedings of
  the 2019 Conference of the North American Chapter of the Association for
  Computational Linguistics: Human Language Technologies, Volume 1 (Long and
  Short Papers), 2019, pp.~3823--3833.

\bibitem{zhou2019dual}
Joey~Tianyi Zhou, Hao Zhang, Di~Jin, Hongyuan Zhu, Meng Fang, Rick Siow~Mong
  Goh, and Kenneth Kwok, \emph{Dual adversarial neural transfer for
  low-resource named entity recognition}, Proceedings of the 57th Annual
  Meeting of the Association for Computational Linguistics, 2019,
  pp.~3461--3471.

\bibitem{zhou2018massively}
Zhong Zhou, Matthias Sperber, and Alex Waibel, \emph{Massively parallel
  cross-lingual learning in low-resource target language translation},
  Proceedings of the Third Conference on Machine Translation: Research Papers,
  2018, pp.~232--243.

\bibitem{kocmi2018trivial}
Tom Kocmi and Ond{\v{r}}ej Bojar, \emph{Trivial transfer learning for
  low-resource neural machine translation}, Proceedings of the Third Conference
  on Machine Translation: Research Papers, 2018, pp.~244--252.

\bibitem{chen2018adversarial}
Xilun Chen, Yu~Sun, Ben Athiwaratkun, Claire Cardie, and Kilian Weinberger,
  \emph{Adversarial deep averaging networks for cross-lingual sentiment
  classification}, Transactions of the Association for Computational
  Linguistics \textbf{6} (2018), 557--570.

\bibitem{rashkin2017truth}
Hannah Rashkin, Eunsol Choi, Jin~Yea Jang, Svitlana Volkova, and Yejin Choi,
  \emph{Truth of varying shades: Analyzing language in fake news and political
  fact-checking}, Proceedings of the 2017 conference on empirical methods in
  natural language processing, 2017, pp.~2931--2937.

\bibitem{shu2020fakenewsnet}
Kai Shu, Deepak Mahudeswaran, Suhang Wang, Dongwon Lee, and Huan Liu,
  \emph{Fakenewsnet: A data repository with news content, social context, and
  spatiotemporal information for studying fake news on social media}, Big Data
  \textbf{8} (2020), no.~3, 171--188.

\bibitem{vo2021hierarchical}
Nguyen Vo and Kyumin Lee, \emph{Hierarchical multi-head attentive network for
  evidence-aware fake news detection}, Proceedings of the 16th Conference of
  the European Chapter of the Association for Computational Linguistics: Main
  Volume, 2021, pp.~965--975.

\bibitem{popat2017truth}
Kashyap Popat, Subhabrata Mukherjee, Jannik Str{\"o}tgen, and Gerhard Weikum,
  \emph{Where the truth lies: Explaining the credibility of emerging claims on
  the web and social media}, Proceedings of the 26th International Conference
  on World Wide Web Companion, 2017, pp.~1003--1012.

\bibitem{pennington2014glove}
Jeffrey Pennington, Richard Socher, and Christopher~D Manning, \emph{Glove:
  Global vectors for word representation}, Proceedings of the 2014 conference
  on empirical methods in natural language processing (EMNLP), 2014,
  pp.~1532--1543.

\bibitem{grave2019unsupervised}
Edouard Grave, Armand Joulin, and Quentin Berthet, \emph{Unsupervised alignment
  of embeddings with wasserstein procrustes}, The 22nd International Conference
  on Artificial Intelligence and Statistics, PMLR, 2019, pp.~1880--1890.

\bibitem{lample2018word}
Guillaume Lample, Alexis Conneau, Marc'Aurelio Ranzato, Ludovic Denoyer, and
  Herv{\'e} J{\'e}gou, \emph{Word translation without parallel data},
  International Conference on Learning Representations, 2018.

\bibitem{kim-2014-convolutional}
Yoon Kim, \emph{Convolutional neural networks for sentence classification},
  Proceedings of the 2014 Conference on Empirical Methods in Natural Language
  Processing ({EMNLP}) (Doha, Qatar), Association for Computational
  Linguistics, October 2014, pp.~1746--1751.

\bibitem{conneau2017very}
Alexis Conneau, Holger Schwenk, Yann Le~Cun, and L{\"o}c Barrault, \emph{Very
  deep convolutional networks for text classification}, 15th Conference of the
  European Chapter of the Association for Computational Linguistics, EACL 2017,
  Association for Computational Linguistics (ACL), 2017, pp.~1107--1116.

\bibitem{martin2017wasserstein}
SC~Martin~Arjovsky and Leon Bottou, \emph{Wasserstein generative adversarial
  networks}, Proceedings of the 34 th International Conference on Machine
  Learning, Sydney, Australia, 2017.

\bibitem{kingma2015adam}
Diederik~P Kingma and Jimmy Ba, \emph{Adam: A method for stochastic
  optimization}, ICLR (Poster), 2015.

\bibitem{luong2015effective}
Minh-Thang Luong, Hieu Pham, and Christopher~D Manning, \emph{Effective
  approaches to attention-based neural machine translation}, Proceedings of the
  2015 Conference on Empirical Methods in Natural Language Processing, 2015,
  pp.~1412--1421.

\bibitem{ganin2016domain}
Yaroslav Ganin, Evgeniya Ustinova, Hana Ajakan, Pascal Germain, Hugo
  Larochelle, Fran{\c{c}}ois Laviolette, Mario Marchand, and Victor Lempitsky,
  \emph{Domain-adversarial training of neural networks}, The journal of machine
  learning research \textbf{17} (2016), no.~1, 2096--2030.

\end{thebibliography}

\section*{Author Biography}

\begin{biography}{\includegraphics[width=60pt,height=70pt]{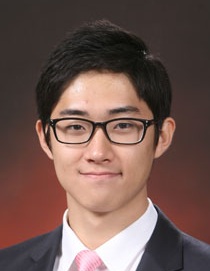}}{\textbf{Sangdo Han.} received his B.S. and M.S. degree at the Department of Computer Science and Engineering at POSTECH. His research interests include dialog system, reading comprehension, and fake news detection.}
\end{biography}


\end{document}